\documentclass[sigconf,authordraft]{acmart}

\AtBeginDocument{%
  \providecommand\BibTeX{{%
    \normalfont B\kern-0.5em{\scshape i\kern-0.25em b}\kern-0.8em\TeX}}}

\copyrightyear{2020}
\acmYear{2020}
\setcopyright{acmlicensed}\acmConference[WebSci '20]{12th ACM Conference on Web Science}{July 6--10, 2020}{Southampton, United Kingdom}
\acmPrice{15.00}
\acmDOI{10.1145/3394231.3397900}
\acmISBN{978-1-4503-7989-2/20/07}

\begin{document}

\title{Gender Classification and Bias Mitigation in Facial Images}

\author{Wenying Wu}
\email{wwu1@alumni.harvard.edu}
\affiliation{%
  \institution{Senior Software Engineer @ Visa Inc.}
}

\author{Pavlos Protopapas}
\email{pavlos@seas.harvard.edu}
\affiliation{%
  \institution{Scientific Program Director @ Harvard John A. Paulson School Of Engineering And Applied Sciences}
}

\author{Zheng Yang}
\email{zhengyang@alumni.harvard.edu}
\affiliation{%
  \institution{Data Scientist @ FM: Systems}
}

\author{Panagiotis Michalatos}
\email{pan.michalatos@gmail.com}
\affiliation{%
  \institution{Senior Principal Research Engineer @ Autodesk}
}

\renewcommand{\shortauthors}{Wu, Michalatos, Protopapas and Yang.}

\begin{abstract}
Gender classification algorithms have important applications in many domains today such as demographic research, law enforcement, as well as human-computer interaction. Recent research showed that algorithms trained on biased benchmark databases could result in algorithmic bias. However, to date, little research has been carried out on gender classification algorithms' bias towards gender minorities subgroups, such as the LGBTQ and the non-binary population, who have distinct characteristics in gender expression. In this paper, we began by conducting surveys on existing benchmark databases for facial recognition and gender classification tasks. We discovered that the current benchmark databases lack representation of gender minority subgroups. We worked on extending the current binary gender classifier to include a non-binary gender class. We did that by assembling two new facial image databases: 1) a racially balanced inclusive database with a subset of LGBTQ population 2) an inclusive-gender database that consists of people with non-binary gender. We worked to increase classification accuracy and mitigate algorithmic biases on our baseline model trained on the augmented benchmark database. Our ensemble model has achieved an overall accuracy score of 90.39\%, which is a 38.72\% increase from the baseline binary gender classifier trained on Adience. While this is an initial attempt towards mitigating bias in gender classification, more work is needed in modeling gender as a continuum by assembling more inclusive databases.
\end{abstract}

\maketitle
\section{Introduction}
\subsection{Gender Classification Algorithm and Its Applications}

Automated gender classification systems recognize the gender of a person based on their facial characteristics that differentiate between masculinity and femininity. Embedded gender classification algorithms in computer software systems have a wide arrange of applications in areas such as video surveillance, law enforcement, demographic research, online advertising, and human-computer interaction \cite{b1}.

Gender information falls under the category of soft biometrics data. Currently, in video surveillance systems, facial soft biometrics features are being extracted to achieve gender and ethnicity classification to aid in security, and forensic evidence collection \cite{b2}. When identifying a target of interest, gender identification could act as a pre-possessing step to reduce the search time in a large-scale database.

In social media design, classifying users' gender is considered as part of the digital monetization effort. Users' demographic data is often times layered on top of coveted behavioral data as part of the market segmentation strategy. For example, gender is redefined through the lens of consumption logic motivated by advertising and revenue generation purposes \cite{b3}. Advertisements that activate identification with one's gender group are more likely to have a favorable impact on future brand and ad judgments  \cite{b4}. 
In HCI today, a variety of intelligent software systems are designed to identify a human's gender attributes to appear more human-like \cite{b5}. For example, when a bot interacts with a human, it would require some information about the subject's gender in order to address them appropriately. 

\subsection{Algorithms Bias in Gender Classification}
As intelligent gender detection and facial recognition systems start to infiltrate into our daily life, ethics in machine learning algorithms has raised significant attention in recent years.  Many machine learning algorithms today can discriminate based on classes such as gender and race -- discriminatory decisions could be made even if the computing process is well-intention-ed.  

Recent studies have found substantial disparities in the accuracy rate of classifying gender of dark-skin females in existing commercial gender classification systems \cite{b6}. Using dermatologist approved Fitzpatrick skin type classification system, dark-skinned females appears to be the most misclassified subclass, with error rates as high as 34.7\% \cite{b6}.

In facial recognition tasks used by the law enforcement systems, research across 100 police departments revealed that African-American individuals are more likely to be subjected to law enforcement searches than individuals of other ethnicity \cite{b7}. The accuracy rates of facial recognition systems used by law enforcement are systematically lower for people labeled female and black \cite{b8}. 

Recent research also discovered that machine learning models are likely to further amplify existing societal gender bias \cite{b9}. For instance, one machine learning model trained for object classification and visual semantic labeling displayed a predictable gender bias in associating activities with binary gender classifier. The current algorithm is 33\% more likely to predict females to be involved in a cooking activity than male \cite{b9}.

\subsection{Unrepresentative Benchmark Databases}
Biased database is a major contributing factor in accuracy disparities among prediction tasks. As the trained model exclusively optimizes accuracy rates on the majority label, samples from the minority class are often times ignored. Most benchmark databases are built using facial detection algorithms to first detect faces from online platforms \cite{b10}. However, any systematic errors in these face detectors will inevitably introduce bias into the benchmark databases. Many databases curated using this approach contain gender and racial bias and lack demographic diversity \cite{b6}. 

For example, Color FERET is a 8.5 GB database released under National Institute of Standards and Technology (NIST) as a benchmark standards for facial recognition under controlled environments \cite{b11}. Color FERET contains very limited number of dark-skinned individuals \cite{b6} \cite{b11}. Even though the database claims to be heterogeneous in its racial and gender distribution, the number of African male identities is only 10\% of their white male counterparts and the number of African female identities is only 17\% of their white female counterparts. 

Labeled Faces in the Wild (LFW), a database composed of celebrity facial images designed for facial verification using the Viola-Jones face detector \cite{b10}, is estimated to be 77.5\% male and 83.5\% white \cite{b12}. Recent facial recognition systems reported a 97.35\% accuracy on the LFW database, however it is not clear how the algorithm is performing on the under-represented subgroups inside the benchmark \cite{b6} \cite{b13}.

\subsection{Benchmark Databases in the Fairness Domain}
Recently, a few benchmark databases have been assembled for facial diversity. For instance, Fairface\cite{b14} is a novel facial image dataset with 108K images curated to mitigate racial bias in datasets.  Collected from YFCC-100M Flickr database, Fairface has a balanced representation of 7 race groups: White, Black, Indian, East Asian, Southeast Asian, Middle East and Latino. Models trained on Fairface dataset reported higher accuracy rates on novel test datasets and the accuracy rates appear to be consistent among different racial and gender subgroups\cite{b14}.

PPB \cite{b6} is a database assembled to achieve better intersectional representation on the basis of gender and skin type. The database consists of 1270 unique identities of government officials and represents a balanced composition of light-skinned individuals and dark-skinned individuals. The PPB database is annotated using the Fitzpatrick skin type labels \cite{b6}. 

Diversity in Faces (DiF) \cite{b15} is another dataset sampled from YFCC-100M Flickr database. DiF contains one million publicly available facial images. The dataset is annotated with various statistical analysis of facial coding scheme measures such as craniofacial features, facial symmetry, facial contrast, skin color, pose, resolution and etc. DiF sets an example for providing a much needed methodology for evaluating diversity in benchmark facial image databases\cite{b15}.

\begin{figure*}[!htb]
  \centering
  \includegraphics[width=\textwidth,height=5cm]{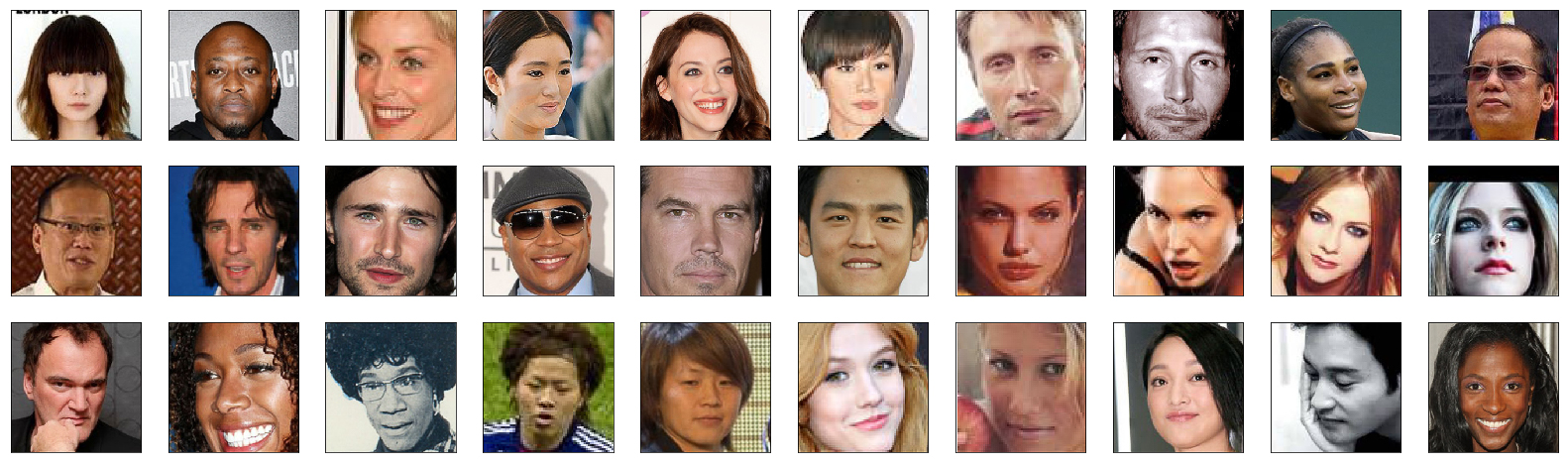}
  \caption{Samples of images from inclusive benchmark dataset}
\end{figure*}

\subsection{The LGBTQ Population and Their Distinct Characteristics in Gender Expression}

In 2017, a controversial study was conducted by Wang, and Kosinski \cite{b16}, who used deep neural networks to detect white males' sexuality. The controversial research implied that facial images of the LGBTQ population have distinct characteristics from the heterosexual population -- LGBTQ population tends to have atypical gender expression and grooming styles \cite{b16}.

A more recent study \cite{b17} released in 2016 estimate the proportion of Americans who identify as transgender to be between 0.5\% and 0.6\%. Transgender facial recognition still remains a challenging area today because gender transformation results in significant face variations over time, both in shape and texture. Gender affirming surgeries often affect the face distribution due to hormone replacement therapy, and depending upon the type of the gender transformation, the changes vary and are subtle to notice. 

The lack of representation of the LGBTQ population in current benchmark databases leads to a false sense of universal progress on gender classification and facial recognition tasks in machine learning. Many current Automatic Gender Recognition (AGR) systems consistently depict gender in a trans-exclusive way. The existence of these systems further reinforces the erasure of the transgender population within our society \cite{b18}. 

\subsection{Limitations of Binary Gender Classification Systems and The Social Distress of Being Misgendered}
The binary gender classification system has created many disadvantages for the sexuality minorities and for those who fall outside the binary gender categories. The presumption that gender is binary, stagnant, and based on physiology, additionally harms the non-binary gendered population.

% With society's belief about strength, weakness and manliness, the binary understanding of gender poses threats on men's mental health well-being. Study reported that men are more likely to suffer from depression and stress from work but less likely to ask for help. Research carried out by the Campaign Against Living Miserably (CALM) and the Huffington Post UK highlighted how men struggled differently to women in mental health issues. The research found that men are more likely to feel pressure to be the main earner in families and they lack the language to talk about their mental health problems. A total of 55\% of men felt males were stereotyped negatively in the media. The survey results reflected the power gender stereotypes that had ingrained from an early age \cite{b10}.
Misgendering refers to the act of using the language to describe somebody that does not align with their affirmed gender identities. A recent study showed that being misgendered can have negative consequences on one's  self-confidence and overall mental health \cite{b19}. 

Gender expression refers to the the way we communicate our gender to others through physical appearances and behaviors such as clothing, hairstyles, and mannerism. The LGBTQ population, whose social gender behavior may not conform to the normative gender roles, are likely to be misgendered. Being misgendered is a form of mockery, imposing a literal challenge to the cis-gender person's stated gender and increasing one's perception of being socially marginalized.

% Currently, an estimate of 0.06\% of global population identify as transgender. 
Sensitivity to transgender people's gender identity is crucially important too because misgendering could cause 'structural violence', potentially turning into catalyst of harassment and discrimination. Misgendering transgender people could result in decreased self-worth and self-esteem, increased level of dysphoria, which could take many in this community to experience depression and even suicidal thoughts \cite{b20}.

Another recent study \cite{b17} also showed that being misgendered by AGR (Automatic Gender Recognition) is worse than being misgendered by human beings. Unlike being misgendered by human beings, AGR systems do not allow users to correct gender classification errors \cite{b21}. Being misgendered by AGR could lead to a greater insult due to perceived objectivity and finitivity of computer systems. Being misgendered by AGR is a reinforcement of gendered standards and another source of invalidation to injure those that were interpreted the wrong gender label.

\section{Methods}
Our benchmark database survey from section 1 revealed that current benchmark databases lack representation of dark-skin individuals, and contain almost no information on the percentage of the LGBTQ population, whose gender expressions often times deviate from the standard gender norms. We hypothesize that by assembling a racially-balanced, LGBTQ and non-binary gender inclusive database, we will be able to improve gender classification algorithms' accuracy on minority faces. We plan to make our assembled non-binary database publicly available for other researchers via \url{https://github.com/trendysloth/nonbinary_gender_benchmark}.

\subsection{Creation of Inclusive Benchmark Database}
We created an inclusive benchmark database with images from different online platforms to achieve better representation of dark-skin individuals and LGBTQ individuals. Our current inclusive database contains 12,000 images of 168 unique identities. Our database contains racially diverse profiles from different geographic locations as well as 21 unique LGBTQ identities which make up for 9\% of the database. Our inclusive database contains 29 white male identities, 25 white female identities, 23 Asian male identities, 23 Asian female identities, 33 African male identities, and 35 African female identities. Figure 1 shows a subset of our inclusive benchmark database.

% \begin{figure}[!ht]
%   \centering
%   \includegraphics[width=\linewidth]{inclusive_identity_distribution.png}
%   \caption{Inclusive Benchmark Database Identity Distribution}
% \end{figure}
% \begin{table}
%   \begin{tabular}{ccccc}
%     \toprule
%     Gender/Race & White & Asian & African & Sum by Gender \\
%     \midrule
%     Male & 29 & 23 & 33 & 85 \\
%     Female & 25 & 23 & 35 & 83 \\
%     Sum by Race & 54 & 46 & 68 & 168 \\
%     \bottomrule
%   \end{tabular}
%   \caption{Inclusive Benchmark Database Identity Distribution}
% \end{table}

\begin{figure}[!htb]
\minipage{0.15\textwidth}
  \includegraphics[width=\linewidth]{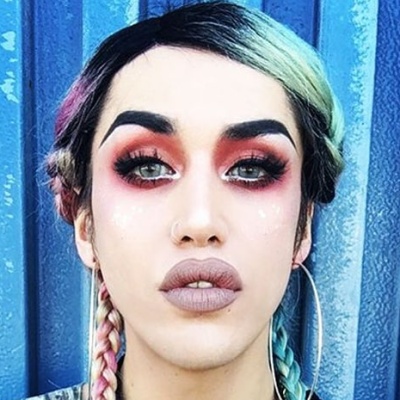}
  \caption{Adore Delano: non-binary}\label{fig:awesome_image1}
\endminipage\hfill
\minipage{0.15\textwidth}
  \includegraphics[width=\linewidth]{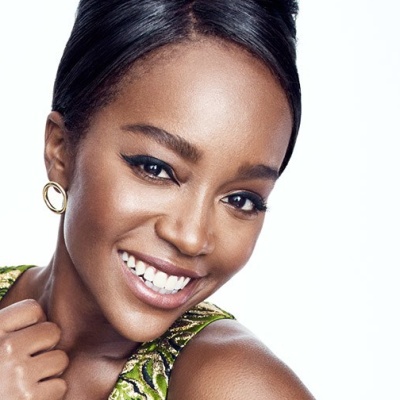}
  \caption{Aja: genderfluid}\label{fig:awesome_image2}
\endminipage\hfill
\minipage{0.15\textwidth}%
  \includegraphics[width=\linewidth]{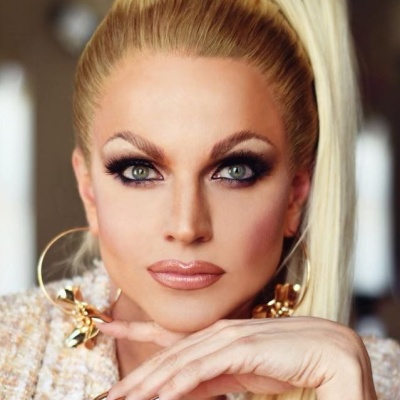}
  \caption{Courtney Act: genderqueer}\label{fig:awesome_image3}
\endminipage
\end{figure}

\begin{table*}[!htb]
\centering
  \begin{tabular}{|c|c|c|c|c|c|c|c|c|}
    \hline
    Database & Source & \# of Images & In-the-wild & \multicolumn{4}{c|}{Gender} & \multicolumn{1}{c|}{Race} \\
    \cline{5-9}
     & & & & M & F & N & Balanced & Balanced\\
    \hline
    ColorFeret & Lab photos & 14K &  & \checkmark & \checkmark &  &  &  \\
    \hline
    LFW & Internet & 13K & \checkmark & \checkmark & \checkmark & & &\\
    \hline
    Adience & Flickr & 18K & \checkmark & \checkmark & \checkmark & & & \\
    \hline
    PPB & Gov. Official & 1K & & \checkmark & \checkmark & & \checkmark & ** skin-type balanced \\
    \hline
    FairFace & Flickr, Twitter, Web & 108K & \checkmark & \checkmark & \checkmark & & \checkmark & \checkmark \\
    \hline
    Inclusive + Non-Binary & Web & 14K & \checkmark & \checkmark & \checkmark &   \checkmark &  \checkmark &  \checkmark \\
    \hline
  \end{tabular}
  \caption{Facial Database Attributes}
\end{table*}

\subsection{Creation of Non-binary Gender Benchmark Database}
To extend the binary gender classifier, we assembled a non-binary gender benchmark database. We decided to use a list of people with non-binary gender identities on Wikipedia since they are public figures with known identities \cite{b22}. Our current Non-binary database include gender identities such as agender, bigender, genderqueer and genderfluid. Our assembled datasets contains 2000 images of 67 unique identities. Figure 2, Figure 3 and Figure 4 show samples of images from our Non-binary Gender Benchmark database. Table 1 shows the gender identity distribution of our Non-binary benchmark database. Table 2 shows a comparison of the annotated facial attributes of our assembled inclusive + non-binary database and other benchmark databases.

\begin{table}
  \begin{tabular}{ccc}
    \hline
    Gender Identity & Counts & Percentage \\
    \hline
    Non-binary & 24 & 35.82\% \\
    Genderfluid & 16 & 23.88\% \\
    Genderqueer & 11 & 17.91\% \\
    Gender non-conforming & 5 & 7.46\% \\
    Agender & 4 & 5.97\% \\
    Gender neutral & 3 & 4.48\% \\
    Genderless & 2 & 3.00\% \\
    Third gender & 1 & 1.50\% \\
    Queer & 1 & 1.50\% \\
    \hline
  \end{tabular}
  \caption{Non-binary Gender Benchmark Database Gender Identity Distribution}
\end{table}

\begin{figure*}[!htb]
  \centering
  \includegraphics[width=\textwidth,height=5cm]{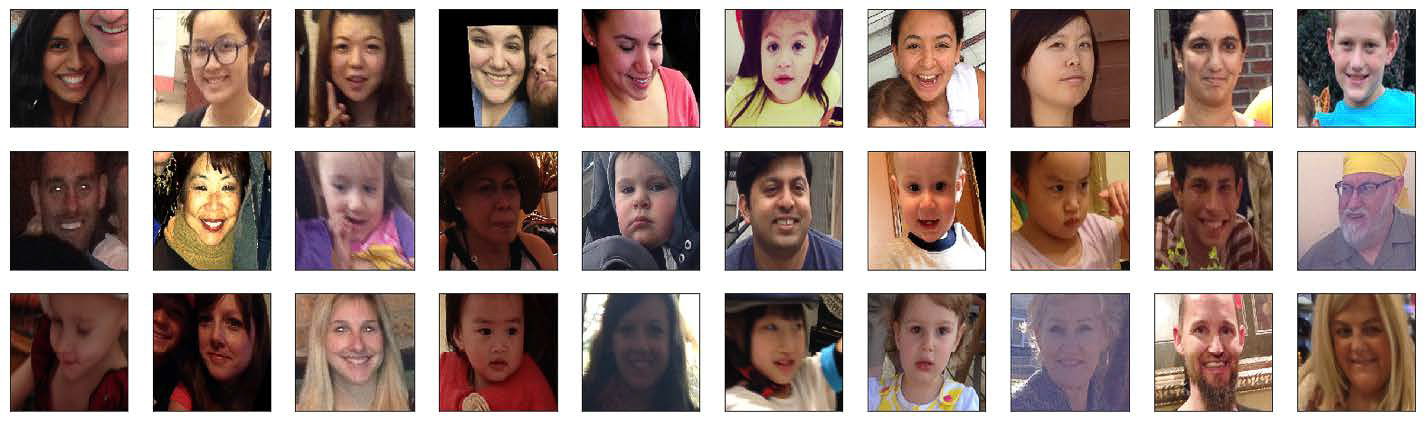}
  \caption{Samples of images from Adience benchmark dataset}
\end{figure*}

\subsection{The Baseline Model Trained on a Biased Benchmark Database}
% As our analysis of gender expression in facial images indicate: there exist distinct profiles for gender expression for the LGBTQ population. However, surveying several large-scale benchmark datasets revealed that currently, there is no information on the percentage of LGBTQ population in any of these benchmark datasets.
Adience is a benchmark facial image database assembled for machine learning research tasks on age and gender classification \cite{b23}. The database represents some of the challenges of age and gender estimation in real-world applications using unconstrained images, such as low-resolution, occlusions, and out-of-plane variations and expressions. The Adience database contains images of individuals of various appearance and postures, with different levels of noise and under different lighting conditions. The current dataset contains 18,270 images, with 2,284 unique subjects from eight age groups (0-2, 4-6, 8-13, 15-20, 25-32, 38-43, 48-53, 60+).

Largely collected through online photography sharing platforms such as Flickr, Adience was highly skewed in its racial representation due to systematic failures on facial detection of dark-skin individuals. Among the 2284 unique identities of the Adience database, only 302 of these identities are dark skinned subjects. Dark-skinned people are highly underrepresented in the Adience benchmark database. The following figures (Figure 5 and Figure 6) show the eigenfaces and a couple of samples from the Adience database.

\begin{figure}[!htb]
  \centering
  \includegraphics[width=\linewidth]{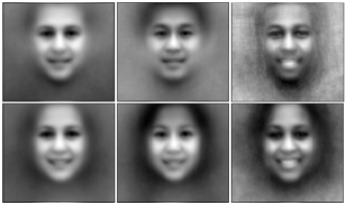}
  \caption{Eigenfaces of Adience Database}
\end{figure}

\begin{figure}[!htb]
  \centering
  \includegraphics[width=\linewidth]{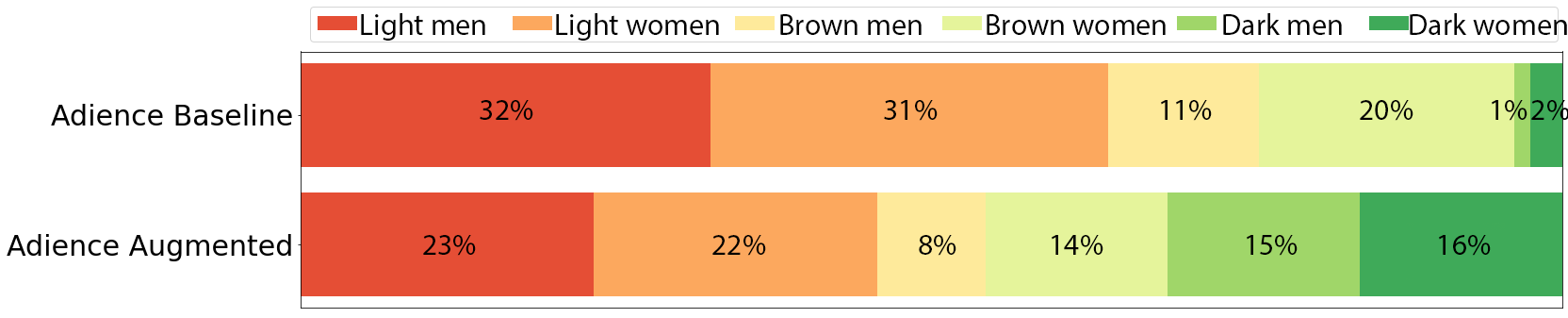}
  \caption{Adience gender, skin-type distribution before and after Data Augmentation}
\end{figure}

\subsubsection{Labeling Process (Adience Benchmark Database)} Gender labels were already annotated in the Adience benchmark dataset. For skin-type labels, one author labeled each image with three categories (Light, Brown, Dark) based on Fitzpatrick skin type (Light skin: Fitzpatrick type 1, type 2; Brown skin: Fitzpatrick type 3, type 4; Dark skin: Fitzpatrick type 5, type 6). The author verified the labels with another person.

\begin{figure*}[!htb]
  \centering
  \includegraphics[width=\textwidth,height=5cm]{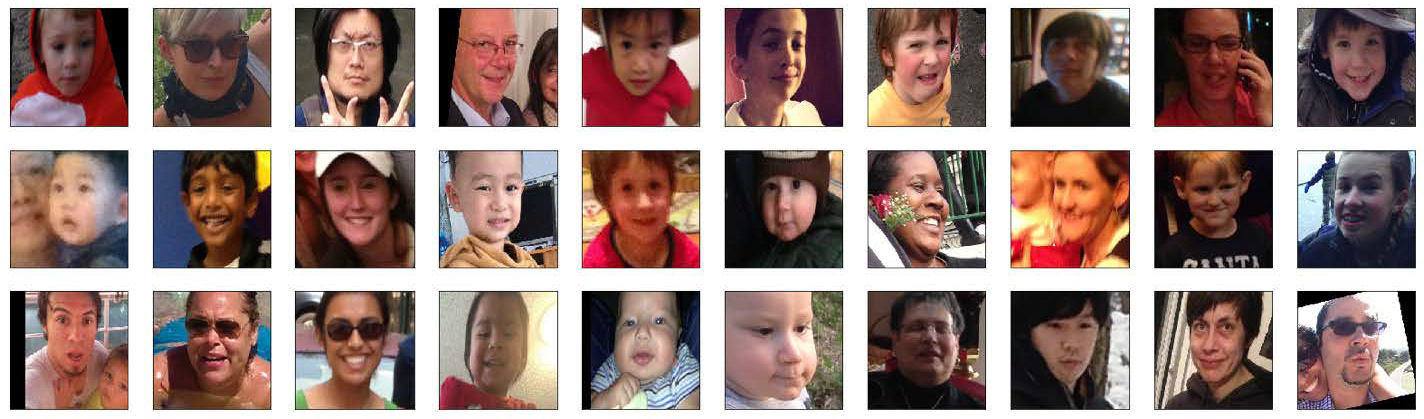}
  \caption{Samples of misclassified images from baseline model}
\end{figure*}

\subsubsection{Data Augmentation}
Given the racially imbalanced nature of the existing Adience benchmark database, we consider several data augmentation techniques as compensatory approaches to artificially inflate the database. Data augmentation will help create a larger and more diverse database without having to collect additional data. By applying data augmentation techniques, we were able to balance the skin-type composition of the database as well as reducing the model's potential risks of overfitting.

The 'ImageDataGenerator' library in Keras offers one of the most effective functions that can be applied to augment data and then use the augmented data for additional training. The way 'ImageDataGenerator' operates is that it computes the internal data stats related to the data dependent transformations \cite{b24}.

Through tuning key parameters and customizing 'ImageDataGenerator' library, we were able to increase the percentage of dark-skin male from 1.3\% to 15.21\% and dark-skin female from 2.5\% to 16.03\%. Figure 7 shows the skin type and gender distribution before and after data augmentation.

% \subsubsection{Data Normalization}
% Considered as a form of data normalization, facial alignment tools helps improve machine learning models' accuracy rates. Recent reports have suggested that face alignment process can substantially boost the performance of face recognition systems and deep learning models. In our project, we applied face alignment by identifying the geometric structure of faces in digital images. We attempted a canonical alignment of the face based on its translation, scale and rotation.

\subsubsection{Training Baseline Model}
After resizing each image to 227 x 227 pixels, we split the augmented Adience benchmark dataset into 23,004 samples for training, 1,279 samples for validation and 1,278 samples for testing.

Inspired by the CNN trained on Adience benchmark database, we trained our baseline model on augmented Adience images using 6 convolutional layers, each followed by a rectified linear operation and pooling layer. The first four layers are also followed by batch normalization and dropouts to prevent the model from overfitting. The first two convolutional layers contain 96 7x7 kernels, the third and fourth convolutional layers contain 256 5x5 kernels. The fifth and final convolutional layers contain 384 3x3 kernels. Additionally, two fully-connected layers are added, each containing 512 neuron.

\subsubsection{Baseline Model Accuracy Evaluation}
Our baseline model is trained to predict gender on two classes: male and female. Our model achieved an overall accuracy score of 94.37\% on the Adience benchmark test set (1278 images). Figure 8 shows a sample of the misclassified images from the baseline and Figure 9 shows the learning curve of our baseline model. Our classifier performs very well in classifying gender for the binary population, except for children, whose secondary gender characteristics aren't fully manifested.

\begin{figure}[!htb]
  \centering
  \includegraphics[width=\linewidth]{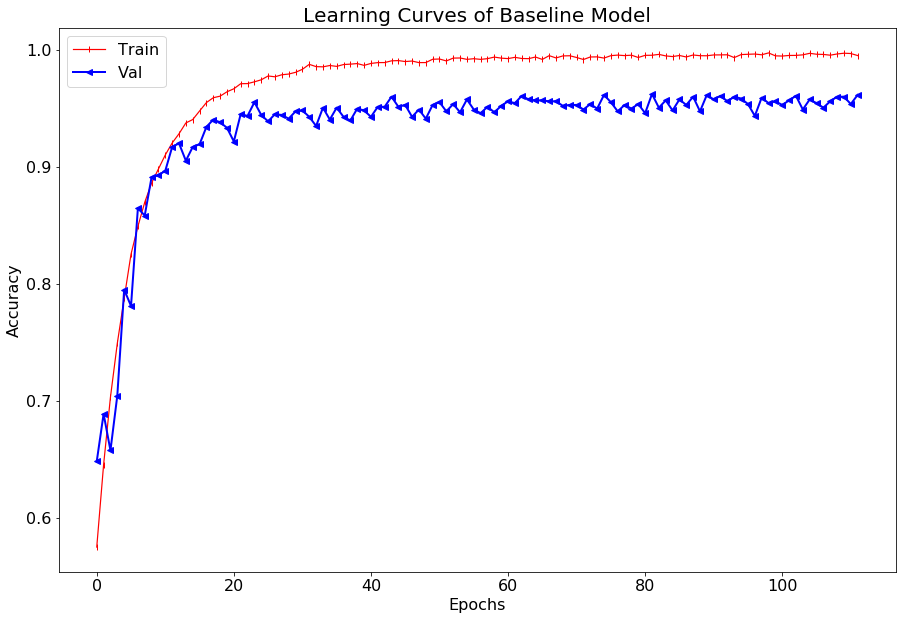}
  \caption{Learning Curves of Baseline Model}
\end{figure}

\subsection{Extension of binary gender classifier}
Using the aforementioned baseline model to predict gender on our test set assembled by inclusive database and non-binary database (3000 images), the overall accuracy dropped to 51.67\%. Our baseline model failed to predict gender on non-binary people whose physical presentation do not fit into the stereotypical male and female categories \cite{b25}.

The identities of non-binary people were rendered as neither legible nor authentic under the baseline's binary system. To extend the current binary gender classifier, we supply training samples from our assembled databases in order to improve neural network's classification performance on non-binary classes.

To extend the binary gender classifier, we supplied 1500 images from the inclusive database, and 1019 images from the non-binary database for training. Similarly, we used 750 images from the inclusive database, and 509 images from the non-binary database for validation, and 750 images from the inclusive database, and 510 images from the non-binary database for testing.

\subsubsection{Oversampling}
Using over-sampling technique, we were able to increase the percentage of male population from 29.89\% (753 images) to 36.51\% (1019 images). Our over-sampled training set contains 2791 images in total. Figure 10 shows the gender distribution before and after sampling.

\begin{figure}[!h]
  \centering
  \includegraphics[width=\linewidth]{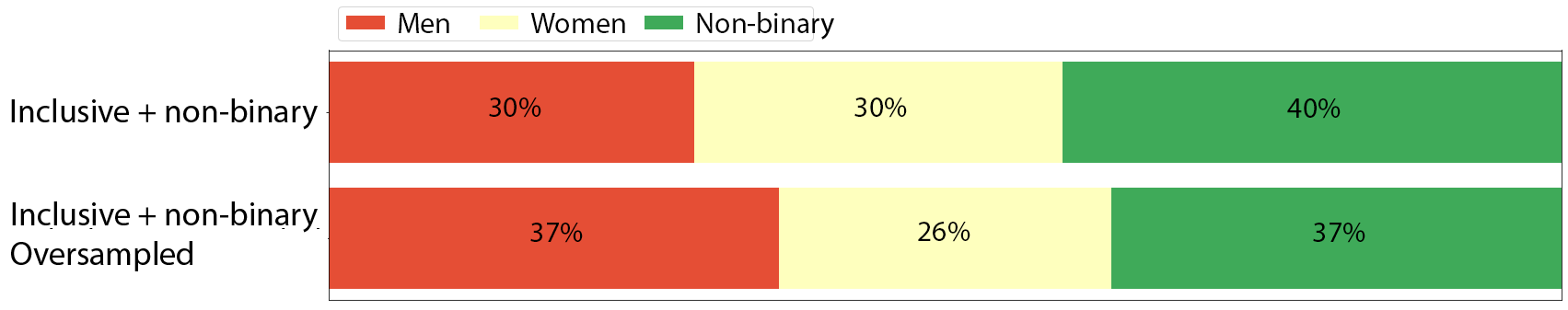}
  \caption{Inclusive + Non-binary gender distribution before and after oversampling}
\end{figure}

\subsection{Transfer Learning Models}
Despite recent successes in machine learning, most machine learning models need a large database and a long time for training. Transfer learning approves to be a useful technique when we are given a new data set that has a different distribution from the original data set. Below, we apply several different transfer learning techniques on our baseline model to improve its classification accuracy on our assembled databases.

\subsubsection{Baseline Feature extraction model}
We freeze the bottom 6 layers of the baseline model trained on the augmented Adience benchmark dataset and then apply transfer learning by popping the last two layers of the baseline and training an additional softmax layer using our oversampled dataset. The overall accuracy for the baseline feature extraction model on our assembled test set (1260 images) increased to 88.10\%. Figure 11 shows the learning curve of baseline feature extraction model.

\begin{figure}[!h]
  \centering
  \includegraphics[width=\linewidth]{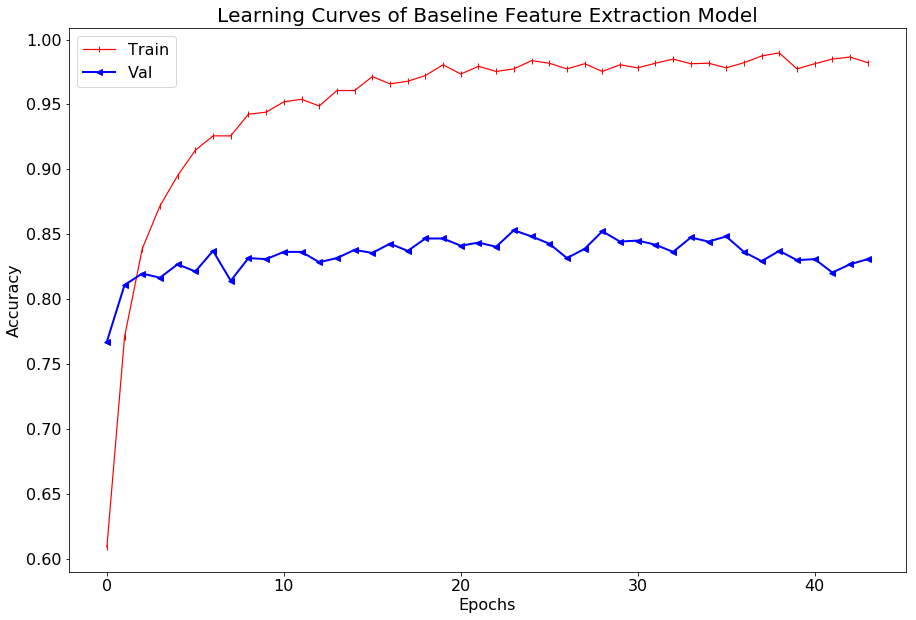}
  \caption{Learning Curves of Baseline Feature Extraction Model}
\end{figure}

\subsubsection{Baseline Fine-tuning Model}
We freeze the top 5 layers of the baseline model trained on augmented Adience benchmark dataset and then apply transfer learning by adding 4 additional convolutional layers on top of the model. The first two convolutional layers contain 64 3x3 kernels. The other two convolutional layers contain 128 3x3 kernels. Both these two blocks of convolutional layers are followed by max pooling and batch normalization. The overall accuracy for the fine-tuned baseline model on the inclusive test set (1260 images) increased to 88.17\%. Figure 12 shows the learning curve of baseline fine-tuning model.

\begin{figure}[!h]
  \centering
  \includegraphics[width=\linewidth]{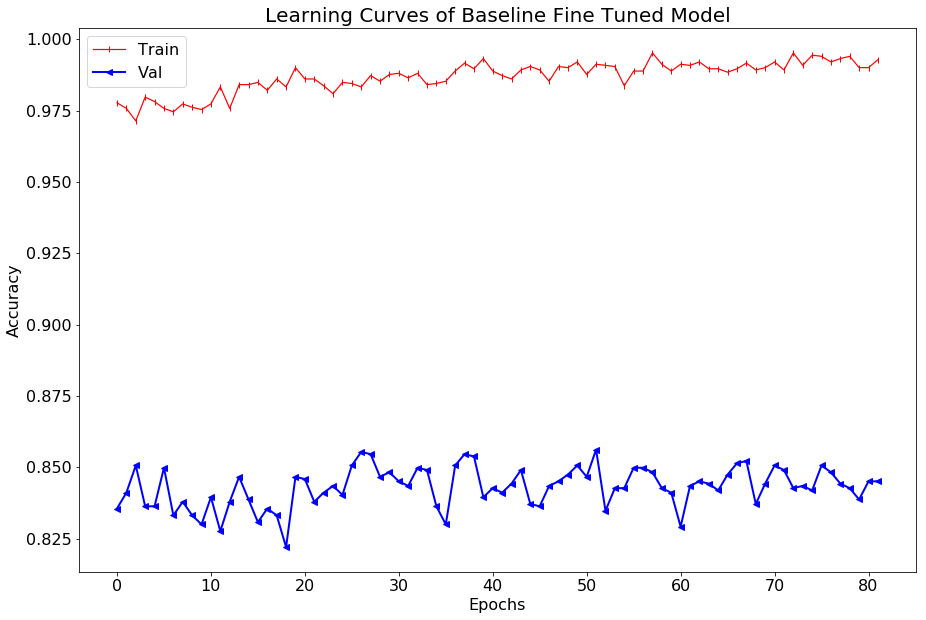}
  \caption{Learning Curves of Baseline Fine-Tuning Model}
\end{figure}

\subsubsection{VGG16 Feature Extraction Model}
The VGG16 architecture contains several convolutional and max pooling layers. The output layer after softmax activation followed by the fully connected layers has 1000 nodes. To apply transfer learning, we freeze all layers of the VGG16 model, pop the last layer of the model and then train an additional softmax layer using our assembled database. The overall accuracy for the VGG16 feature extraction model on the assembled test set (1260 images) is 85.32\%. Figure 13 shows the learning curve of VGG16 feature extraction model.

\begin{figure}[!h]
  \centering
  \includegraphics[width=\linewidth]{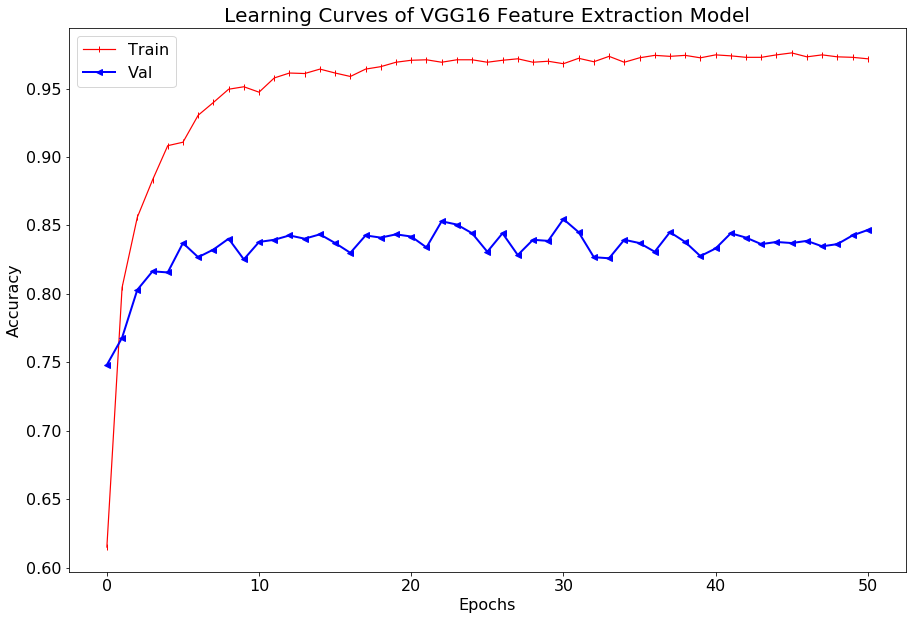}
  \caption{Learning Curves of VGG16 Feature Extraction Model}
\end{figure}

Our various learning curves figures above show that the accuracy rates of our training sets are around 10\% higher than that of our validation set. This implied that our models are overfitted. In order to reduce the effect of overfitting, we worked with various ensemble techniques below to make our final prediction outcomes more robust.

\subsubsection{Logistic Regression Ensemble}
We constructed a logistic regression model by assembling the baseline fine-tuned model, the baseline feature extraction model and the VGG16 feature extraction model  using the logistic regression class. By horizontally stacking the probabilities of the three classes output by the list of models, we constructed a matrix of dimension 3778 x 9 which we used for training. After applying logistic regression cross validation, we trained a logistic regression model and achieved an accuracy score of 96.03 \% on the training set. Applying the trained model on the test set (1260 images), we received an accuracy score of 90.39\%. 

\begin{table*}[!htb]
\centering
  \begin{tabular}{|c|c|c|c|c|c|c|}
    \hline
    Model & Wrong Images & \multicolumn{4}{c|}{Accuracy Rate} & Selection Rate \\
    \cline{3-6}
     & & Overall & Male & Female & Non-binary & \\
    \hline
    Baseline & 609 & 51.67\% & 87.87\% & 85.75\% & 0\% & 0\% \\
    \hline
    Baseline Feature Extraction & 150 & 88.10\% & 89.76\% & 85.49\% & 88.82\% & 95.24\% \\
    \hline
    Basline Fine-tuned & 149 & 88.17\% & 84.10\% & 88.39\% & 90.98\% & 92.44\% \\
    \hline
    VGG16 Feature Extraction & 185 & 85.32\% & 85.71\% & 83.38\% & 86.47\% & 96.43\% \\
    \hline
    Logistic Regression Ensemble & 121 & 90.39\% & 90.02\% & 88.65\% & 91.97\% & 96.40\% \\
    \hline
    Adaboost Ensemble & 123 & 90.24\% & 91.11\% & 89.71\% & 90.00\% & 98.46\% \\
    \hline
  \end{tabular}
  \caption{Model Accuracy Rates and Selection Rates}
\end{table*}

\subsubsection{Adaboost Ensemble}
We constructed an adaboost model by assembling the baseline fine-tuned model, the baseline feature extraction model and the VGG16 feature extraction model using the an adaboost classifier class. By horizontally stacking the predicted class output by each of the models, we constructed a matrix of dimension of 3778 x 3 which we used for training. After applying grid search to cross validate the parameters, we trained an adaboost mega-model and achieved an accuracy score of 96.03\%. Applying the trained model on the test set (1260 images), we received an accuracy score of 90.24\%.  

\begin{figure*}[!htb]
  \centering
  \includegraphics[width=\linewidth,height=5cm]{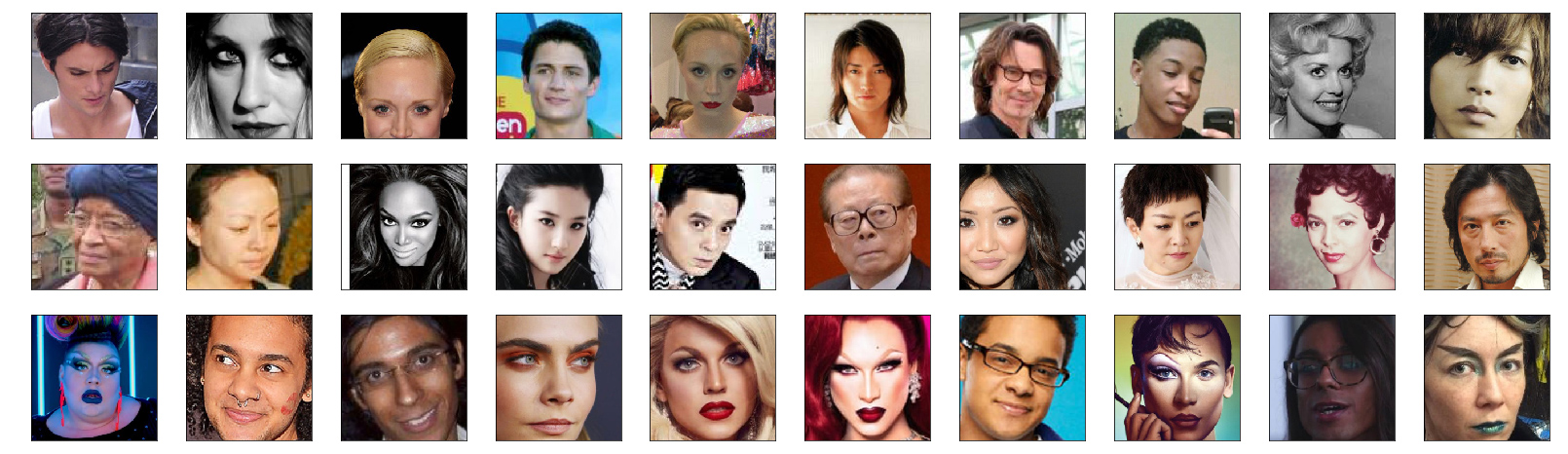}
  \caption{Samples of misclassified images using logistic ensemble model}
\end{figure*}

\subsection{Evaluation Metric}
\begin{equation} \label{eq1}
\begin{split}
selection\, rate &= \frac{1 - error\, of\, worst}{1 - error\, of\, best}\\
&= \frac{worst\, accuracy\, rate}{best\, accuracy\, rate}
\end{split}
\end{equation}

We use selection rate \cite{b26} as a measure of algorithmic bias. An increase in selection rate is considered to be bias mitigation in which case the differences between the highest accuracy rate and the lowest accuracy rate has been reduced.

Applying the 80\% threshold, we consider any selection rate below 80\% as an indication of disparate impact. Disparate impact can result in a disparate proportion of individuals from a protected group (e.g., race, gender, religion, etc.) being subjected to unintended discrimination \cite{b26}. 

Using the 80\% rule, our baseline model is subject to disparate impact since the accuracy rate for non-binary gender population is 0\%. After applying transfer learning techniques however, disparate impact is removed. Transfer learning has produced impressive results in improving classification accuracy on the non-binary gender class and mitigating algorithmic biases. Our Adaboost ensemble model achieved a selection rate of 98.46\%, a significant increase from our baseline model. These results demonstrated the importance of assembling inclusive databases on machine learning tasks to enable accurate labeling and fairness in decision making. Table 3 shows the accuracy rates breakdown and the selection rate of our various machine learning models.

\section{Results and Discussion}
We trained a binary gender classifier after augmenting a racially-imbalanced facial image database, Adience. We then measured the accuracy of gender classification on our assembled inclusive database and our non-binary gender database. We found that our baseline model is subject to disparate impact on the non-binary population under the 80\% rule.

After applying various machine learning techniques such as oversampling, transfer learning and ensemble, we were able to mitigate algorithmic bias and increase classification accuracy from our baseline model. We assembled three of our transfer learning models: baseline feature extraction, baseline fine-tuned, and VGG16 feature extraction model using logistic ensemble -- our mega model achieved an overall accuracy rate of 90.39\%.

As results shown in Table 3, when compared to the baseline model trained on Adience benchmark database, our improved gender classification algorithm has the following advantages:
\begin{itemize}
  \item Our classifier has extended the binary gender classifier to predict gender on the non-binary class.
  \item Our classifier has achieved good accuracy rates in predicting gender for the non-binary population. Our best performing model (logistic regression ensemble) has achieved an accuracy score of 91.97\%. 
  \item Our algorithm has mitigated algorithmic bias from the baseline. Our adaboost ensemble model has achieved a selection rate of 98.46\%.
\end{itemize}

\section{Limitations and future work}
\subsection{Limitations}

\begin{figure*}[!h]
  \centering
  \includegraphics[width=\linewidth,height=5cm]{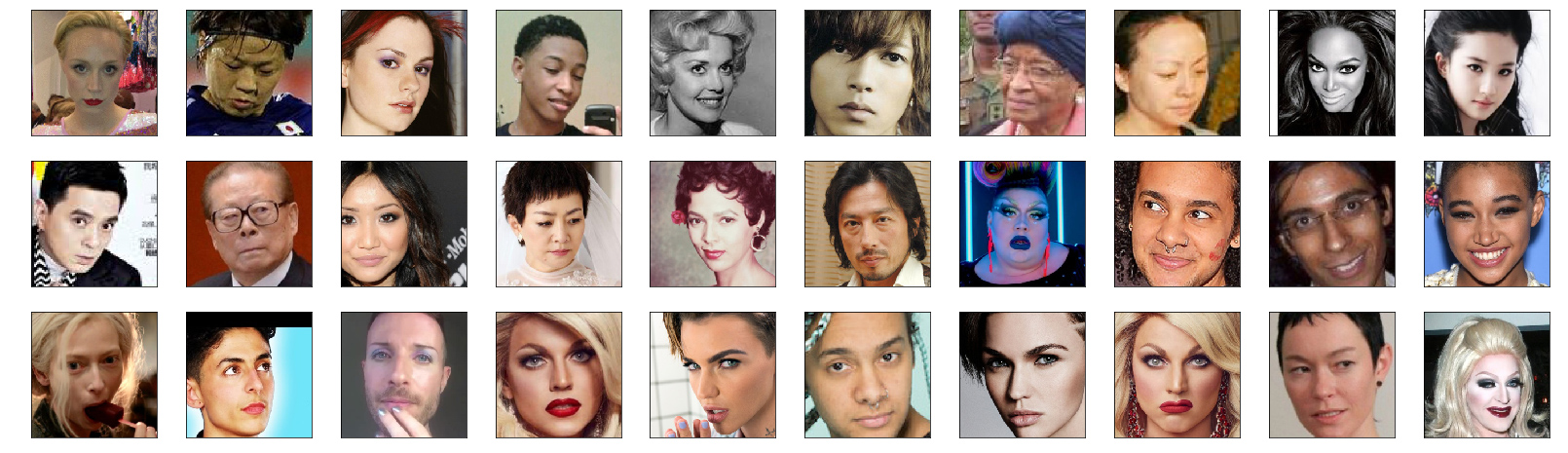}
  \caption{Samples of misclassified images using adaboost ensemble model}
\end{figure*}

Our study extended the current binary gender classification system to include a non-binary gender class. By assembling 2 different databases (a racially balanced and LGBTQ inclusive database and a non-binary gender database), we worked with various machine learning techniques to teach the classifier to recognize non-binary gender class and increase gender classification accuracy for the non-binary population. However, our algorithm is limited in terms of providing specific gender labels within the non-binary class, such as gender fluid, gender-queer, gender nonconforming, agender, demigender and etc.

Among the misclassified images using our best performing classifier -- logistic regression ensemble model, we noticed that people of color, as well as the LGBTQ population are more likely to be misclassified compared to the other subclasses. Figure 14 shows a sample of misclassified images using our logistic ensemble model. And Figure 15 shows a sample of the misclassified images using our adaboost ensemble model. The classification accuracy is also affected by different masks on face. For example, we noticed that people who are wearing glasses, or heavy makeups are more likely to be misclassified. Increasing the classification accuracy on these minority subclasses implies the importance of assembling more inclusive databases for facial recognition machine learning tasks. 

\subsection{Future Work: Modeling Gender as a Continuum}
In her 1988 essay 'Performative Acts and Gender Constitution', Judith Butler defined gender as performative actions associated with male or female. Gender identity is constructed through the repeated stylization of the body within a regulatory social framework that in turn reinforces the existing social constructions of gender \cite{b27}. Our concept of gender is seen as natural or innate because the body becomes its gender through a set of actions which are renewed, revised, and consolidated through time \cite{b28}. 

Gender is a complex socio-cultural construct and an internal identity that is not necessarily tied to physical appearances. Gender identity has its multifaceted aspects that a simple label could not categorize. For instance, even when the algorithm accurately identifies a trans woman as a woman, there are many layers of details specific to transgender experiences and social contexts that such a shallow label will miss \cite{b17}.

Gender expressions also vary in different social contexts. A recent study found that non-binary people are inclined to respond to others' perception of their gender entering different social contexts. Some people shift their gender expression to appear more feminine or masculine in order to fit into the social context \cite{b29}. Some may find their non-binary gender identity as relatively private so that they do not express themselves in the public space.

According to a recent survey by the LGBTQ advocacy organization GLAAD, around 20\% of the millennials identified themselves other than strictly cis-gendered \cite{b30}. Gender expressions that go beyond the fixed labeling terms are moving from the margins to the mainstream. On that note, our curated non-binary database is not evident in representing those who reject the notion of having a gender identity label in the first place. Apart from this, our algorithm does not recognize gender identities that overlaps between semantic labels such as agender/genderless and genderqueer/genderfluid.

Building on top of our current database and gender detection algorithms, our future work will embark on modeling gender identity as a continuum (with 'femininity' and 'masculinity' as two endpoints). By hypothesizing that secondary characteristics are strong indicators of gender, we are looking to quantify gender expression by annotating facial images using secondary characteristics measures such as the roll, pitch, yaw angle of a human head. 

Using this approach, we hope to break the static gender labeling system, and represent gender identities by measuring gender expressions using quantitative values. The gender continuum model will therefore allow more fluidity and variation in gender expressions in different social contexts, and be especially beneficial to those whose gender identities do not comply with binary assumptions.

\begin{acks}
We thank Joy Buolamwini from MIT for providing us with Fitzpatrick skin type labels on Adience benchmark database. 
\end{acks}

\end{document}